# A MODIFIED VORTEX SEARCH ALGORITHM FOR NUMERICAL FUNCTION OPTIMIZATION

Berat Doğan

Department of Biomedical Engineering, Inonu University, Malatya, Turkey

**ABSTRACT**

*The Vortex Search (VS) algorithm is one of the recently proposed metaheuristic algorithms which was inspired from the vortical flow of the stirred fluids. Although the VS algorithm is shown to be a good candidate for the solution of certain optimization problems, it also has some drawbacks. In the VS algorithm, candidate solutions are generated around the current best solution by using a Gaussian distribution at each iteration pass. This provides simplicity to the algorithm but it also leads to some problems along. Especially, for the functions those have a number of local minimum points, to select a single point to generate candidate solutions leads the algorithm to being trapped into a local minimum point. Due to the adaptive step-size adjustment scheme used in the VS algorithm, the locality of the created candidate solutions is increased at each iteration pass. Therefore, if the algorithm cannot escape a local point as quickly as possible, it becomes much more difficult for the algorithm to escape from that point in the latter iterations. In this study, a modified Vortex Search algorithm (MVS) is proposed to overcome above mentioned drawback of the existing VS algorithm. In the MVS algorithm, the candidate solutions are generated around a number of points at each iteration pass. Computational results showed that with the help of this modification the global search ability of the existing VS algorithm is improved and the MVS algorithm outperformed the existing VS algorithm, PSO2011 and ABC algorithms for the benchmark numerical function set.*

## 1. INTRODUCTION

In the past two decades, a number of metaheuristic algorithms have been proposed to solve complex real-world optimization problems. Most of these algorithms are nature inspired methods and therefore mimic natural metaphors such as, evolution of species (GA [1] and DE [2-3]), annealing process (SA [4-5]), ant behaviour (ACO [6]), swarm behaviour (PSO [7] and ABC [8-9]) etc. These algorithms make few or no assumptions for the problem at hand and provide fast and robust solutions. Although, the solutions provided by metaheuristics may not be optimal solutions, they are highly preferred because of their simplicity and flexibility.

Despite the high number of available metaheuristics, developing new metaheuristic algorithms is still an active research area. In [10-15], a number of recently proposed metaheuristics can be found. All of these metaheuristics have certain characteristics and thus each one may be more successful on a certain optimization problem when compared to the others. The Vortex Search (VS) algorithm [16] is one of these recently proposed metaheuristic algorithms which was inspired from the vortical flow of the stirred fluids. The search behaviour of the VS algorithm is modelled as a vortex pattern by using an adaptive step-size adjustment scheme. By this way, it is aimed to have a good balance between the explorative and exploitative behaviour of the search.

The proposed VS algorithm was tested over 50 benchmark mathematical functions and the obtained results compared to the single-solution based (Simulated Annealing, SA and Pattern Search, PS) and population-based (Particle Swarm Optimization, PSO2011 and Artificial Bee Colony, ABC) algorithms. A Wilcoxon-Signed Rank Test was performed to measure the pair-wise statistical performances of the algorithms, the results of which indicated that the proposed VS algorithm outperforms the SA, PS and ABC algorithms while being competitive with the PSO2011 algorithm. Because of the simplicity of the proposed VS algorithm, a significant decrease in the computational time of the 50 benchmark numerical functions was also achieved when compared to the population-based algorithms. In some other studies [17-20], the VS algorithm has also been successfully used for the solution of some real-world optimization problems.

Although the proposed VS algorithm is a good candidate for the solution of optimization problems, it also has some drawbacks. In the VS algorithm, candidate solutions are generated around the current best solution by using a Gaussian distribution at each iteration pass. This provides simplicity to the algorithm but it also leads to some problems along. Especially, for the functions those have a number of local minimum points, to select a single point to generate candidate solutions leads the algorithm to being trapped into a local minimum point. Due to the adaptive step-size adjustment scheme used in the VS algorithm, the locality of the created candidate solutions is increased at each iteration pass. Therefore, if the algorithm cannot escape a local point as quickly as possible, it becomes much more difficult for the algorithm to escape from that point in the latter iterations.

In this study, a modified Vortex Search algorithm (MVS) is proposed to overcome above mentioned drawback of the existing VS algorithm. In the MVS algorithm, the candidate solutions are generated around different points at each iteration pass. These points are iteratively updated during the search process, details of which are given in the following section. The MVS algorithm is again tested by using the 50 benchmark mathematical functions that was used earlier in [16]. Because the SA and PS algorithms showed poor performances in [16], in this study these two algorithms are excluded and the results are compared to the results those obtained by the VS algorithm, PSO2011 and ABC algorithms. It is shown that, the MVS algorithm outperforms all of these algorithms and can successfully find the known global minimum points of the functions that the VS algorithm being trapped into the local minimum points earlier.

The remaining part of this paper is organized as follows. In the following section, first a brief description of the VS algorithm is given. Then, the modification performed on the VS algorithm is detailed and the MVS algorithm is introduced. Section 3 covers the experimental results and discussion. Finally, Section 4 concludes the work.

## 2. METHODOLOGY

### 2.1. A Brief Description of the Vortex Search Algorithm

Let us consider a two-dimensional optimization problem. In a two dimensional space a vortex pattern can be modelled by a number of nested circles. Here, the outer (largest) circle of the vortex is first centered on the search space, where the initial center can be calculated using Eq. 1.

$$\mu_0 = \frac{upperlimit + lowerlimit}{2} \qquad (1)$$

In Eq.1, *upperlimit* and *lowerlimit* are $d \times 1$ vectors that define the bound constraints of the problem in $d$ dimensional space. Then, a number of neighbor solutions $C_t(s)$, ($t$ represents the iteration index and initially $t = 0$) are randomly generated around the initial center $\mu_0$ in the $d$-dimensional space by using a Gaussian distribution. Here, $C_0(s) = \{s_1, s_2, ..., s_k\}$ $k = 1, 2, ..., n$ represents the solutions, and $n$ represents the total number of candidate solutions. In Eq. 2, the general form of the multivariate Gaussian distribution is given.

$$p(x | \mu, \Sigma) = \frac{1}{\sqrt{(2\pi)^d |\Sigma|}} \exp\left\{-\frac{1}{2}(x-\mu)^T \Sigma^{-1}(x-\mu)\right\} \qquad (2)$$

In Eq.2, $d$ represents the dimension, $x$ is the $d \times 1$ vector of a random variable, $\mu$ is the $d \times 1$ vector of sample mean (center) and $\Sigma$ is the covariance matrix. If the diagonal elements (variances) of the values of $\Sigma$ are equal and if the off-diagonal elements (covariance) are zero (uncorrelated), then the resulting shape of the distribution will be spherical (which can be considered circular for a two-dimensional problem, as in our case). Thus, the value of $\Sigma$ can be computed by using equal variances with zero covariance by using Eq. 3.

$$\Sigma = \sigma^2 \cdot [I]_{d \times d} \qquad (3)$$

In Eq. 3, $\sigma^2$ represents the variance of the distribution and $I$ represents the $d \times d$ identity matrix. The initial standard deviation ($\sigma_0$) of the distribution can be calculated by using Eq. 4.

$$\sigma_0 = \frac{\max(upperlimit) - \min(lowerlimit)}{2} \qquad (4)$$

Here, $\sigma_0$ can also be considered as the initial radius ($r_0$) of the outer circle for a two dimensional optimization problem. Because a weak locality is required in the initial phases, $r_0$ is chosen to be a large value. Thus, a full coverage of the search space by the outer circle is provided in the initial step. This process provides a bird's-eye view for the problem at hand.

In the selection phase, a solution (which is the best one) $s' \in C_0(s)$ is selected and memorized from $C_0(s)$ to replace the current circle center $\mu_0$. Prior to the selection phase, the candidate solutions must be ensured to be inside the search boundaries. For this purpose, the solutions that exceed the boundaries are shifted into the boundaries, as in Eq. 5.

$$s_k^i = \begin{cases} rand \cdot (upperlimit^i - lowerlimit^i) + lowerlimit^i, & s_k^i < lowerlimit^i \\ s_k^i, & lowerlimit^i \leq s_k^i \leq upperlimit^i \\ rand \cdot (upperlimit^i - lowerlimit^i) + lowerlimit^i, & s_k^i > upperlimit^i \end{cases} \qquad (5)$$

In Eq.5, $k = 1, 2, ..., n$ and $i = 1, 2, ..., d$ and $rand$ is a uniformly distributed random number. Next, the memorized best solution $s'$ is assigned to be the center of the second circle (the inner one). In the generation phase of the second step, the effective radius ($r_1$) of this new circle is reduced, and then, a new set of solutions $C_1(s)$ is generated around the new center. Note that in the second step, the locality of the generated neighbors increased with the decreased radius. In the selection phase of the second step, the new set of solutions $C_1(s)$ is evaluated to select a solution $s' \in C_1(s)$. If the selected solution is better than the best solution found so far, then this solution is assigned to be the new best solution and it is memorized. Next, the center of the third circle is

International Journal of Artificial Intelligence and Applications (IJAIA), Vol. 7, No. 3, May 2016

assigned to be the memorized best solution found so far. This process iterates until the termination condition is met. An illustrative sketch of the process is given in Figure 1. In this manner, once the algorithm is terminated, the resulting pattern appears as a vortex-like structure, where the center of the smallest circle is the optimum point found by the algorithm. A representative pattern is sketched in Figure 2 for a two-dimensional optimization problem for which the upper and lower limits are between the [-10,10] interval. A description of the VS algorithm is also provided in Figure 3.

The radius decrement process given in Figure 3 can be considered as a type of adaptive step-size adjustment process which has critical importance on the performance of the VS algorithm. This process should be performed in such a way that allows the algorithm to behave in an explorative manner in the initial steps and in an exploitative manner in the latter steps. To achieve this type of process, the value of the radius must be tuned properly during the search process. In the VS algorithm, the inverse incomplete gamma function is used to decrease the value of the radius during each iteration pass.

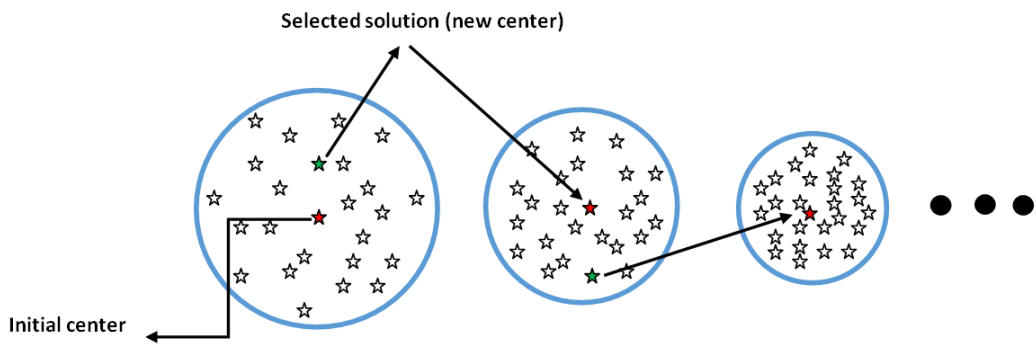

Figure 1. An illustrative sketch of the search process

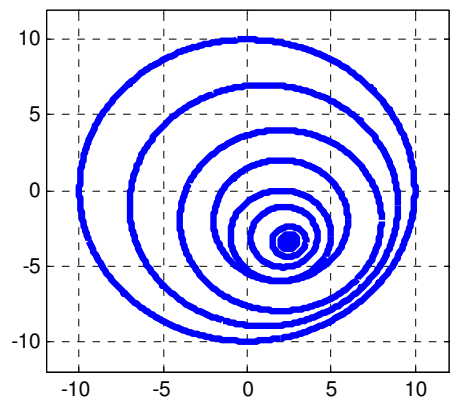

Figure 2. A representative pattern showing the search boundaries (circles) of the VS algorithm after a search process, which has a vortex-like structure.

The incomplete gamma function given in Eq. 6 most commonly arises in probability theory, particularly in those applications involving the chi-square distribution [21].

$$\gamma(x,a) = \int_0^x e^{-t} t^{a-1} dt \quad a > 0 \tag{6}$$

In Eq.6, $a > 0$ is known as the shape parameter and $x \geq 0$ is a random variable. In conjunction with the incomplete gamma function, its complementary $\Gamma(x,a)$ is usually also introduced (Eq. 7).

$$\Gamma(x,a) = \int_x^\infty e^{-t} t^{a-1} dt \quad a > 0 \tag{7}$$

Thus, it follows that,

$$\gamma(x,a) + \Gamma(x,a) = \Gamma(a) \tag{8}$$

where $\Gamma(a)$ is known as the gamma function. There exist many studies in the literature on different proposed methods for the numerical calculation of the incomplete gamma function [22-24]. MATLAB® also provides some tools for the calculation of the inverse incomplete gamma (*gammaincinv*) function. The inverse incomplete gamma function (*gammaincinv*), computes the inverse of the incomplete gamma function with respect to the integration limit $x$ and represented as *gammaincinv(x,a)* in MATLAB®.

---

**Inputs:** Initial center $\mu_0$ is calculated by using Eq. 1
Initial radius $r_0$ (or the standard deviation, $\sigma_0$) is computed by using Eq. 10
Fitness of the best solution found so far $f(s_{best}) = \inf$
$t = 0$;
**Repeat**
/* Generate candidate solutions by using Gaussian distribution around the center $\mu_t$ with a standard deviation (radius) $r_t$*/
Generate($C_t(s)$);
If exceeded, then shift the $C_t(s)$ values into the boundaries as in Eq.5
/* Select the best solution from $C_t(s)$ to replace the current center $\mu_t$ */
$s' = $ Select($C_t(s)$);
if $f(s') < f(s_{best})$
$\quad s_{best} = s'$
$\quad f(s_{best}) = f(s')$
else
$\quad$ keep the best solution found so far $s_{best}$
**end**
/* Center is always shifted to the best solution found so far */
$\mu_{t+1} = s_{best}$
/* Decrease the standard deviation (radius) for the next iteration */
$r_{t+1} = $ Decrease($r_t$)
$t = t + 1$;
**Until** the maximum number of iterations is reached
**Output:** Best solution found so far $s_{best}$

---

Figure 3. A description of the VS algorithm

In Figure 4, the inverse incomplete gamma function is plotted for $x = 0.1$ and $a \in [0,1]$. Here, for our case the parameter $a$ of the inverse incomplete gamma function defines the resolution of the search. By equally sampling $a$ values within $[0,1]$ interval at a certain step size, the resolution of the search can be adjusted. For this purpose, at each iteration, a value of $a$ is computed by using the Eq.9

$$a_t = a_0 - \frac{t}{MaxItr} \qquad (9)$$

where $a_0$ is selected as $a_0 = 1$ to ensure a full coverage of the search space at the first iteration, $t$ is the iteration index, and $MaxItr$ represents the maximum number of iterations.

Let us consider an optimization problem defined within the [-10,10] region. The initial radius $r_0$ can be calculated with Eq. 10. Because $a_0 = 1$, the resulting function value is $(1/x) \cdot gammaincinv(x, a_0) \approx 1$, which means $r_0 \approx \sigma_0$ as indicated before.

$$r_0 = \sigma_0 \cdot (1/x) \cdot gammaincinv(x, a_0) \qquad (10)$$

By means of Eq.4, the initial radius value $r_0$ can be calculated as $r_0 \approx 10$. In Eq.11, a general formula is also given to obtain the value of the radius at each iteration pass.

$$r_t = \sigma_0 \cdot (1/x) \cdot gammaincinv(x, a_t) \qquad (11)$$

Here, $t$ represents the iteration index.

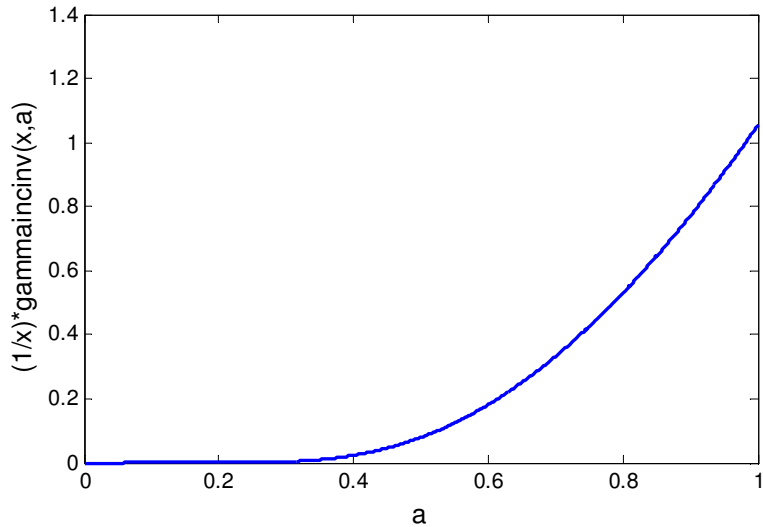

Figure 4. $(1/x) \cdot gammaincin\, v(x, a)$ where $x = 0.1$ and $a \in [0,1]$

## 2.2. The Modified Vortex Search Algorithm

The VS algorithm creates candidate solutions around a single point at each iteration pass. At the first iteration, this point is the initial center $\mu_0$ which is determined with the upper and lower limits of the problem at hand while in the latter iterations the center is shifted to the current best position found so far. As mentioned before, this mechanism leads the VS algorithm to being trapped into local minimum points for a number of functions.

To overcome above mentioned drawback, in this study a modified VS algorithm (MVS) is proposed. In the MVS algorithm, candidate solutions are generated around multiple centers at each iterations pass. The search behavior of the MVS algorithm can be thought as a number of parallel vortices that have different centers at each iteration pass. Initially, the centers of these multiple vortices are selected as in the VS algorithm. Let us consider, the total number of centers (or vortices) to be represented by $m$. Let us say, $M_t(\mu)$ represents the matrix that stores the values of these $m$ centers at each iteration pass and $t$ represents the iteration index. Thus, initially $M_0(\mu) = \{\mu_0^1, \mu_0^2, ..., \mu_0^l\}$ $l = 1,2,...,m$ and initial positions of these centers are computed as in Eq. 12.

$$\mu_0^1 = \mu_0^2 = ... = \mu_0^l = \frac{upperlimit + lowerlimit}{2}, \quad l = 1,2,...,m \tag{12}$$

Next, a number of candidate solutions are generated with a Gaussian distribution around these initial centers by using the initial radius value $r_0$. In this case the total number of candidate solutions is again selected to be $n$. But note that, these $n$ solutions are generated around $m$ centers. Thus, one should select $n/m$ solutions around each center.

Let us say, $CS_t^l(s) = \{s_1, s_2, ..., s_k\}$ $k = 1,2,...,n/m$ represents the subset of solutions generated around the center $l = 1,2,...,m$ for the iteration $t$. Then, the total solution set generated for the iteration $t = 0$ can be represented by $C_0(s) = \{CS_0^1, CS_0^2, ..., CS_0^l\}, l = 1,2,...,m$. In the selection phase, for each subset of solutions, a solution (which is the best one) $s_l^{'} \in CS_0^l(s)$ is selected. Prior to the selection phase it must be ensured that the candidate subsets of solutions are inside the search boundaries. For this purpose, the solutions that exceed the boundaries are shifted into the boundaries, as in Eq. 5. Let us say, the best solution of each subset is stored in a matrix $PBest_t(s^{'})$ at each iteration pass. Thus, for $t = 0$, $PBest_0(s^{'}) = \{s_1^{'}, s_2^{'}, ..., s_l^{'}\} l = 1,2,...,m$. Note that, the best solution of this matrix ($PBest_0(s^{'})$) is also the best solution of the total candidate solution set $C_0(s)$ for the current iteration, which is represented as $Itr_{best}$.

In the VS algorithm, at each iterations pass, the center is always shifted to the best solution found so far, $s_{best}$. However, in the MVS algorithm, there exist $m$ centers which positions need to be updated for the next iteration. The most important difference between the VS and MVS algorithm arises from here. In the MVS algorithm, one of these centers is again shifted to the best solution found so far, $s_{best}$. But, the remaining $m-1$ centers are shifted to a new position determined by the best positions generated around the each center at the iteration $t$ and the best position found so far, $s_{best}$ as shown in Eq. 13.

$$\mu_t^l = s_l^{'} + rand \cdot (s_l^{'} + s_{best}) \tag{13}$$

In Eq. 13, $rand$ is a uniformly distributed random number, $l = 1,2,...,m-1$ and $s_l^{'} \in PBest_{t-1}(s^{'})$. Thus, for $t = 1$, $M_1(\mu) = \{\mu_1^1, \mu_1^2, ..., \mu_1^l\}$ $l = 1,2,...,m-1$ is determined by using the $s_l^{'} \in PBest_0(s^{'})$ positions and the best position found so far, $s_{best}$. In Figure-5, an illustrative sketch of the center update process is given for a two-dimensional problem. In Figure 5, only one center is considered.

In the MVS algorithm, the radius decrement process is held totally in the same way as it is done in the VS algorithm. At each iteration pass, the radius is decreased by utilizing the inverse incomplete gamma function and thus, the locality of the generated solutions is increased. In Figure 6, a description of the MVS algorithm is provided.

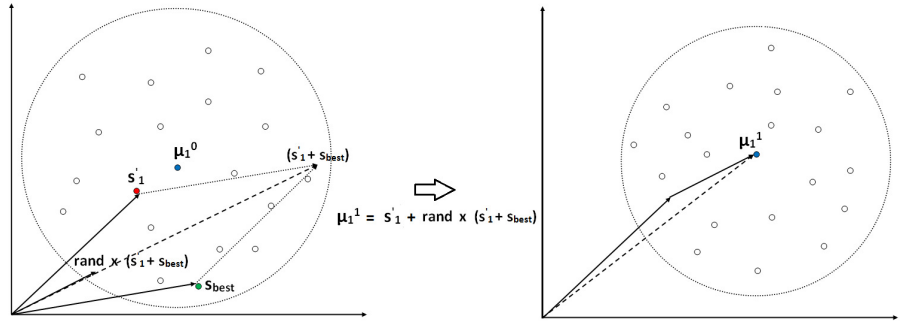

Figure 5. An illustrative sketch of the center updating process for the MVS algorithm (only one center is considered)

**Inputs:** Initial centers $M_0(\mu) = \{\mu_0^1, \mu_0^2, ..., \mu_0^l\}$, $l = 1, 2, ..., m$ is computed by using Eq. 12
Initial radius $r_0$ (or the standard deviation, $\sigma_0$) is computed by using Eq. 10
Fitness of the best solution found so far $f(s_{best}) = \inf$
$t = 0$;
**Repeat**
    /* Generate candidate solution sets by using Gaussian distribution around the centers with a standard deviation (radius) $r_t$ */
    Generate($CS_t^l(s)$);
    If exceeded, then shift the $CS_t^l(s)$ values into the boundaries as in Eq.5
    /* Select the best solution from each subset $CS_t^l(s)$ to update the corresponding centers $\mu_t^l$ */
    $s_l' = \text{Select}(CS_t^l(s))$;
    /* Store the best solution of each subset $CS_t^l(s)$ into the matrix $PBest_t(s')$ */
    $PBest_t(s') = \text{Store}(s_l')$
    /* Select the best solution $Itr_{best}$ from the $PBest_t(s')$ */
    $Itr_{best} = \text{Select}(PBest_t(s'))$
    if $f(Itr_{best}) < f(s_{best})$
        $s_{best} = Itr_{best}$
        $f(s_{best}) = f(Itr_{best})$
    else
        keep the best solution found so far $s_{best}$
    end

    Shift $m-1$ centers to their new positions as in Eq. 13
    Shift one of the centers to the best solution found so far, $s_{best}$

    /* Decrease the standard deviation (radius) for the next iteration */
    $r_{t+1} = \text{Decrease}(r_t)$
    $t = t+1$;
**Until** the maximum number of iterations is reached
**Output:** Best solution found so far $s_{best}$

Figure 6. A description of the MVS algorithm

## 3. RESULTS

The proposed MVS algorithm is tested on 50 benchmark functions which were also used in [16] to measure the performance of the VS algorithm. By using the same functions, in this study, the performance of the MVS algorithm is compared to the VS, PSO2011 and ABC algorithms. PSO2011 [25-26] is an extension of the standard PSO algorithm and the ABC algorithm is a well-known optimization algorithm which was inspired from the collective behaviours of honey bees.

The functions used in the experiments are listed in Table 1. For the formulations of the functions listed in Table 1, please refer to the reference [16].

### 3.1. Algorithm Settings

The ABC and PSO2011 algorithms are selected to have a population size of 50, which is also the number of neighborhood solutions of the proposed VS algorithm. The acceleration coefficients ($c_1$ and $c_2$) of the PSO2011 algorithm are both set to 1.8, and the inertia coefficient is set to 0.6, as in [27]. The *limit* value for the ABC algorithm is determined as *limit = SN * D,* where *SN* represents the number of food sources and *D* represents the dimension. VS algorithm does not have any additional parameters. Different from the VS algorithm, the MVS algorithm has the parameter $m$, which represents the total number of centers.

### 3.2. Results

The proposed MVS algorithm is compared to the VS, PSO2011 and ABC algorithms by using the 50 benchmark functions given in Table 1. For each algorithm, 30 different runs are performed, and the mean and the best values are recorded. The maximum number of iterations is selected to be 500,000. For the MATLAB® codes of the PSO2011, ABC, VS and MVS algorithms please refer to [25], [28], [29] and [30]. For each algorithm, all of the functions are run in parallel using a 32 core Intel® CPU 32 GB RAM workstation. For the first set of experiments, results are given in Table 2.

As shown in Table 2, for the MVS algorithm two different cases are considered. In the first case, the total number of candidate solutions is selected to be 50, which means 10 candidate solutions are generated around each center for $m = 5$. In this case, the MVS algorithm can avoid from the local minimum points for the functions F13, F16, F17, F22, F23, F41, and F43 which is not the case for the VS algorithm. However, poor sampling of the search space for this case (10 points around each center) leads the MVS algorithm to show a correspondingly poor performance on the improvement of the found near optimal solutions (exploitation) for some of the functions. This can be clearly seen in the pair-wise statistical comparison of the algorithms given in Table 3. The pair-wise statistical comparison of the algorithms is obtained by a Wilcoxon Signed-Rank Test. The null hypothesis $H_0$ for this test is: "There is no difference between the median of the solutions produced by algorithm A and the median of the solutions produced by algorithm B for the same benchmark problem", i.e., median (A) = median (B). To determine whether algorithm A reached a statistically better solution than algorithm B, or if not, whether the alternative hypothesis is valid, the sizes of the ranks provided by the Wilcoxon Signed-Rank Test (i.e., T+ and T-, as defined in [10]) are examined. Because an arithmetic precision value that is higher than necessary makes it difficult to compare the local search abilities of the algorithms, during the statistical pair-wise comparison, resulting values below are considered as 0. In Table 3, '+' indicates cases in which the null hypothesis is rejected and the MVS algorithm exhibited a statistically superior performance in the pair-wise Wilcoxon Signed-Rank Test at the 95% significance level ($\alpha = 0.05$); '-' indicates cases in which the null hypothesis is rejected and the MVS algorithm exhibited an inferior performance; and '=' indicates cases in which there is no statistical different between two algorithms. The last row of the Table 3 shows the total count of (+/=/-) for the three statistical significance cases in the pair-wise comparison.

As shown in Table 3, the VS algorithm outperforms the MVS algorithm for the functions F5, F10, F11, F38, and F42. But when the results given in Table 2 are compared for these functions, it can be clearly seen that this difference mainly arises because of the poor exploitation ability of the MVS algorithm with 50 candidate solutions. Therefore, another case in which the total number of candidate solutions is selected to be 250 is considered for the MVS algorithm. In this case, 50 candidate solutions are generated around each center for $m=5$. As can be shown in Table 2, the MVS algorithm with 250 candidate solutions performs better than the MVS algorithm with 50 candidate solutions. Statistical pair-wise comparison of the algorithms for the second case is also given in Table 4.

Table 1. Benchmark function set that is used in the experiments.

| No | Function | Characteristics | Range | Dim. | Min. |
|---|---|---|---|---|---|
| F1 | Stepint | Unimodal Separable | [-5.12, 5.12] | 5 | 0 |
| F2 | Step | Unimodal Separable | [-100, 100] | 30 | 0 |
| F3 | Sphere | Unimodal Separable | [-100, 100] | 30 | 0 |
| F4 | SumSquares | Unimodal Separable | [-10, 10] | 30 | 0 |
| F5 | Quartic | Unimodal Separable | [-1.28, 1.28] | 30 | 0 |
| F6 | Beale | Unimodal Non-Separable | [-4.5, 4.5] | 5 | 0 |
| F7 | Easom | Unimodal Non-Separable | [-100, 100] | 2 | -1 |
| F8 | Matyas | Unimodal Non-Separable | [-10, 10] | 2 | 0 |
| F9 | Colville | Unimodal Non-Separable | [-10, 10] | 4 | 0 |
| F10 | Trid6 | Unimodal Non-Separable | $[-D^2,D^2]$ | 6 | -50 |
| F11 | Trid10 | Unimodal Non-Separable | $[-D^2,D^2]$ | 10 | -210 |
| F12 | Zakharov | Unimodal Non-Separable | [-5,10] | 10 | 0 |
| F13 | Powell | Unimodal Non-Separable | [-4,5] | 24 | 0 |
| F14 | Schwefel 2.22 | Unimodal Non-Separable | [-10, 10] | 30 | 0 |
| F15 | Schwefel 1.2 | Unimodal Non-Separable | [-10, 10] | 30 | 0 |
| F16 | Rosenbrock | Unimodal Non-Separable | [-30, 30] | 30 | 0 |
| F17 | Dixon-Price | Unimodal Non-Separable | [-10, 10] | 30 | 0 |
| F18 | Foxholes | Multimodal Separable | [-65.536, 65.536] | 2 | 0.998 |
| F19 | Branin | Multimodal Separable | [-5,10]x[0,15] | 2 | 0.398 |
| F20 | Bohachevsky1 | Multimodal Separable | [-100, 100] | 2 | 0 |
| F21 | Booth | Multimodal Separable | [-10, 10] | 2 | 0 |
| F22 | Rastrigin | Multimodal Separable | [-5.12, 5.12] | 30 | 0 |
| F23 | Schwefel | Multimodal Separable | [-500, 500] | 30 | -12569.5 |
| F24 | Michalewicz2 | Multimodal Separable | $[0,\pi]$ | 2 | -1.8013 |
| F25 | Michalewicz5 | Multimodal Separable | $[0,\pi]$ | 5 | -4.6877 |
| F26 | Michalewicz10 | Multimodal Separable | $[0,\pi]$ | 10 | -9.6602 |
| F27 | Schaffer | Multimodal Non-Separable | [-100, 100] | 2 | 0 |
| F28 | Six Hump Camel Back | Multimodal Non-Separable | [-5, 5] | 2 | -1.03163 |
| F29 | Bohachevsky2 | Multimodal Non-Separable | [-100, 100] | 2 | 0 |
| F30 | Bohachevsky3 | Multimodal Non-Separable | [-100, 100] | 2 | 0 |
| F31 | Shubert | Multimodal Non-Separable | [-10, 10] | 2 | -186.73 |
| F32 | GoldStein-Price | Multimodal Non-Separable | [-2, 2] | 2 | 3 |
| F33 | Kowalik | Multimodal Non-Separable | [-5, 5] | 4 | 0.00031 |
| F34 | Shekel5 | Multimodal Non-Separable | [0, 10] | 4 | -10.15 |
| F35 | Shekel7 | Multimodal Non-Separable | [0, 10] | 4 | -10.4 |
| F36 | Shekel10 | Multimodal Non-Separable | [0, 10] | 4 | -10.53 |
| F37 | Perm | Multimodal Non-Separable | [D, D] | 4 | 0 |
| F38 | PowerSum | Multimodal Non-Separable | [0, D] | 4 | 0 |
| F39 | Hartman3 | Multimodal Non-Separable | [0, 1] | 3 | -3.86 |
| F40 | Hartman6 | Multimodal Non-Separable | [0, 1] | 6 | -3.32 |
| F41 | Griewank | Multimodal Non-Separable | [-600, 600] | 30 | 0 |

| F42 | Ackley | Multimodal Non-Separable | [-32, 32] | 30 | 0 |
| F43 | Penalized | Multimodal Non-Separable | [-50, 50] | 30 | 0 |
| F44 | Penalized2 | Multimodal Non-Separable | [-50, 50] | 30 | 0 |
| F45 | Langerman2 | Multimodal Non-Separable | [0, 10] | 2 | -1.08 |
| F46 | Langerman5 | Multimodal Non-Separable | [0, 10] | 5 | -1.5 |
| F47 | Langerman10 | Multimodal Non-Separable | [0, 10] | 10 | NA |
| F48 | Fletcher Powell2 | Multimodal Non-Separable | $[-\pi, \pi]$ | 2 | 0 |
| F49 | Fletcher Powell5 | Multimodal Non-Separable | $[-\pi, \pi]$ | 5 | 0 |
| F50 | Fletcher Powell10 | Multimodal Non-Separable | $[-\pi, \pi]$ | 10 | 0 |

Table 2. Statistical results of 30 runs obtained by PSO2011, ABC, VS and MVS algorithms (values < $10^{-16}$ are considered as 0).

| No | Min. | | MVS (m = 5, n = 50) | MVS (m = 5, n = 250) | VS | PSO2011 | ABC |
|---|---|---|---|---|---|---|---|
| F1 | 0 | Mean | 0 | 0 | 0 | 0 | 0 |
| | | StdDev | 0 | 0 | 0 | 0 | 0 |
| | | Best | 0 | 0 | 0 | 0 | 0 |
| F2 | 0 | Mean | 0 | 0 | 0.2 | 0.066666667 | 0 |
| | | StdDev | 0 | 0 | 0.406838102 | 0.253708132 | 0 |
| | | Best | 0 | 0 | 0 | 0 | 0 |
| F3 | 0 | Mean | 0 | 0 | 0 | 0 | 2.78624E-16 |
| | | StdDev | 0 | 0 | 0 | 0 | 0 |
| | | Best | 0 | 0 | 0 | 0 | 2.23487E-16 |
| F4 | 0 | Mean | 0 | 0 | 0 | 0 | 2.75098E-16 |
| | | StdDev | 0 | 0 | 0 | 0 | 0 |
| | | Best | 0 | 0 | 0 | 0 | 1.85594E-16 |
| F5 | 0 | Mean | 0.000334677 | 0.000127606 | 0.000145026 | 1.64098E-05 | 0.013732963 |
| | | StdDev | 0.000316498 | 7.01148E-05 | 7.30549E-05 | 5.56581E-06 | 0.002379448 |
| | | Best | 3.62661E-07 | 5.59873E-06 | 5.54996E-05 | 7.13993E-06 | 0.008413424 |
| F6 | 0 | Mean | 0 | 0 | 0 | 0 | 6.37598E-16 |
| | | StdDev | 0 | 0 | 0 | 0 | 3.58687E-16 |
| | | Best | 0 | 0 | 0 | 0 | 0 |
| F7 | -1 | Mean | -1 | -1 | -1 | -1 | -1 |
| | | StdDev | 0 | 0 | 0 | 0 | 0 |
| | | Best | -1 | -1 | -1 | -1 | -1 |
| F8 | 0 | Mean | 0 | 0 | 0 | 0 | 0 |
| | | StdDev | 0 | 0 | 0 | 0 | 0 |
| | | Best | 0 | 0 | 0 | 0 | 0 |
| F9 | 0 | Mean | 0 | 0 | 0 | 0 | 0.00576453 |
| | | StdDev | 0 | 0 | 0 | 0 | 0.003966867 |
| | | Best | 0 | 0 | 0 | 0 | 0.000383073 |
| F10 | -50 | Mean | -50 | -50 | -50 | -50 | -50 |
| | | StdDev | 2.88473E-14 | 2.97973E-14 | 2.96215E-14 | 3.61345E-14 | 4.94748E-14 |
| | | Best | -50 | -50 | -50 | -50 | -50 |
| F11 | -210 | Mean | -210 | -210 | -210 | -210 | -210 |
| | | StdDev | 6.64246E-13 | 5.51673E-13 | 6.19774E-13 | 2.30778E-13 | 9.62204E-12 |
| | | Best | -210 | -210 | -210 | -210 | -210 |
| F12 | 0 | Mean | 0 | 0 | 0 | 0 | 7.56674E-14 |
| | | StdDev | 0 | 0 | 0 | 0 | 3.76382E-14 |
| | | Best | 0 | 0 | 0 | 0 | 2.31887E-14 |
| F13 | 0 | Mean | 7.59934E-09 | 3.88377E-10 | 1.43967E-05 | 2.04664E-07 | 9.09913E-05 |
| | | StdDev | 4.14437E-08 | 1.27749E-09 | 2.27742E-06 | 1.21051E-08 | 1.42475E-05 |

| | | | | | | |
|---|---|---|---|---|---|---|
| | Best | 1.1432E-16 | 0 | 5.71959E-06 | 1.72679E-07 | 5.23427E-05 |
| F14 0 | Mean | 3.79651E-05 | 0 | 0 | 1.094284383 | 8.51365E-16 |
| | StdDev | 0.000156663 | 0 | 0 | 0.870781136 | 0 |
| | Best | 0 | 0 | 0 | 0.107097937 | 6.93597E-16 |
| F15 0 | Mean | 0 | 0 | 0 | 0 | 0.000760232 |
| | StdDev | 0 | 0 | 0 | 0 | 0.000440926 |
| | Best | 0 | 0 | 0 | 0 | 0.00027179 |
| F16 0 | Mean | 1.20813E-07 | 3.51659E-08 | 0.367860114 | 0.930212233 | 0.003535257 |
| | StdDev | 2.94163E-07 | 5.41004E-08 | 1.130879848 | 1.714978077 | 0.003314818 |
| | Best | 1.14463E-12 | 1.85577E-13 | 9.42587E-05 | 0 | 7.08757E-05 |
| F17 0 | Mean | 0 | 0 | 0.666666667 | 0.666666667 | 1.91607E-15 |
| | StdDev | 0 | 0 | 7.68909E-16 | 4.38309E-16 | 2.55403E-16 |
| | Best | 0 | 0 | 0.666666667 | 0.666666667 | 1.1447E-15 |

Table 2. (continued).

| No | Min. | | MVS (m = 5, n = 50) | MVS (m = 5, n = 250) | VS | PSO2011 | ABC |
|---|---|---|---|---|---|---|---|
| F18 | 0.998 | Mean | 0.998003838 | 0.998003838 | 0.998003838 | 34.26621987 | 0.998003933 |
| | | StdDev | 0 | 0 | 0 | 126.6004794 | 4.33771E-07 |
| | | Best | 0.998003838 | 0.998003838 | 0.998003838 | 0.998003838 | 0.998003838 |
| F19 | 0.398 | Mean | 0.397887358 | 0.397887358 | 0.397887358 | 0.397887358 | 0.397887358 |
| | | StdDev | 0 | 0 | 0 | 0 | 0 |
| | | Best | 0.397887358 | 0.397887358 | 0.397887358 | 0.397887358 | 0.397887358 |
| F20 | 0 | Mean | 0 | 0 | 0 | 0 | 0 |
| | | StdDev | 0 | 0 | 0 | 0 | 0 |
| | | Best | 0 | 0 | 0 | 0 | 0 |
| F21 | 0 | Mean | 0 | 0 | 0 | 0 | 0 |
| | | StdDev | 0 | 0 | 0 | 0 | 0 |
| | | Best | 0 | 0 | 0 | 0 | 0 |
| F22 | 0 | Mean | 4.14483E-16 | 0 | 57.60799224 | 26.11016129 | 0 |
| | | StdDev | 8.95296E-16 | 3.24317E-16 | 13.94980276 | 5.686650032 | 0 |
| | | Best | 0 | 0 | 33.82857771 | 16.91429893 | 0 |
| F23 | -12569.5 | Mean | -12569.48662 | -12569.48662 | -11283.05416 | -8316.185447 | -12569.48662 |
| | | StdDev | 3.63798E-12 | 3.02118E-12 | 352.1869262 | 463.9606712 | 1.85009E-12 |
| | | Best | -12569.48662 | -12569.48662 | -11799.62928 | -9466.201047 | -12569.48662 |
| F24 | -1.8013 | Mean | -1.80130341 | -1.80130341 | -1.80130341 | -1.80130341 | -1.80130341 |
| | | StdDev | 9.03362E-16 | 9.03362E-16 | 9.03362E-16 | 9.03362E-16 | 9.03362E-16 |
| | | Best | -1.80130341 | -1.80130341 | -1.80130341 | -1.80130341 | -1.80130341 |
| F25 | -4.6877 | Mean | -4.653710247 | -4.668168867 | -4.670953055 | -4.67700874 | -4.687658179 |
| | | StdDev | 0.051587389 | 0.02119113 | 0.020809276 | 0.036487971 | 2.60778E-15 |
| | | Best | -4.687658179 | -4.687658179 | -4.687658179 | -4.687658179 | -4.687658179 |
| F26 | -9.6602 | Mean | -8.966488952 | -9.07030728 | -8.793361668 | -9.204154798 | -9.660151716 |
| | | StdDev | 0.412225375 | 0.267235114 | 0.382153549 | 0.298287637 | 0 |
| | | Best | -9.556414106 | -9.513891389 | -9.410563187 | -9.660151716 | -9.660151716 |
| F27 | 0 | Mean | 0 | 0 | 0 | 0 | 0 |
| | | StdDev | 0 | 0 | 0 | 0 | 0 |
| | | Best | 0 | 0 | 0 | 0 | 0 |
| F28 | -1.03163 | Mean | -1.031628453 | -1.031628453 | -1.031628453 | -1.031628453 | -1.031628453 |
| | | StdDev | 6.77522E-16 | 6.77522E-16 | 6.77522E-16 | 6.71219E-16 | 6.77522E-16 |
| | | Best | -1.031628453 | -1.031628453 | -1.031628453 | -1.031628453 | -1.031628453 |
| F29 | 0 | Mean | 0 | 0 | 0 | 0 | 0 |
| | | StdDev | 0 | 0 | 0 | 0 | 0 |
| | | Best | 0 | 0 | 0 | 0 | 0 |
| F30 | 0 | Mean | 0 | 0 | 0 | 0 | 0 |
| | | StdDev | 0 | 0 | 0 | 0 | 0 |
| | | Best | 0 | 0 | 0 | 0 | 0 |

| No | Min. | | | | | | |
|---|---|---|---|---|---|---|---|
| F31 | -186.73 | Mean | -186.7309088 | -186.7309088 | -186.7309088 | -186.7309088 | -186.7309088 |
| | | StdDev | 3.25344E-14 | 2.93854E-14 | 3.76909E-14 | 4.49449E-13 | 1.18015E-14 |
| | | Best | -186.7309088 | -186.7309088 | -186.7309088 | -186.7309088 | -186.7309088 |
| F32 | 3 | Mean | 3 | 3 | 3 | 3 | 3 |
| | | StdDev | 1.25607E-15 | 1.51835E-15 | 1.44961E-15 | 1.22871E-15 | 1.7916E-15 |
| | | Best | 3 | 3 | 3 | 3 | 3 |
| F33 | 0.00031 | Mean | 0.000307486 | 0.000307486 | 0.000307486 | 0.000307486 | 0.000319345 |
| | | StdDev | 0 | 0 | 0 | 0 | 5.4385E-06 |
| | | Best | 0.000307486 | 0.000307486 | 0.000307486 | 0.000307486 | 0.00030894 |
| F34 | -10.15 | Mean | -10.15319968 | -10.15319968 | -10.15319968 | -9.363375596 | -10.15319968 |
| | | StdDev | 6.8481E-15 | 7.01294E-15 | 7.2269E-15 | 2.081063878 | 7.2269E-15 |
| | | Best | -10.15319968 | -10.15319968 | -10.15319968 | -10.15319968 | -10.15319968 |

Table 2. (continued).

| No | Min. | | MVS (m = 5, n = 50) | MVS (m = 5, n = 250) | VS | PSO2011 | ABC |
|---|---|---|---|---|---|---|---|
| F35 | -10.4 | Mean | -10.40294057 | -10.40294057 | -10.40294057 | -10.40294057 | -10.40294057 |
| | | StdDev | 1.51161E-15 | 1.64931E-15 | 1.61598E-15 | 1.80672E-15 | 1.04311E-15 |
| | | Best | -10.40294057 | -10.40294057 | -10.40294057 | -10.40294057 | -10.40294057 |
| F36 | -10.53 | Mean | -10.53640982 | -10.53640982 | -10.53640982 | -10.53640982 | -10.53640982 |
| | | StdDev | 2.05998E-15 | 1.61598E-15 | 1.47518E-15 | 0 | 2.13774E-15 |
| | | Best | -10.53640982 | -10.53640982 | -10.53640982 | -10.53640982 | -10.53640982 |
| F37 | 0 | Mean | 0.003329829 | 0.003422267 | 0.002815467 | 0.002854996 | 0.003526435 |
| | | StdDev | 0.002357345 | 0.002379677 | 0.002374325 | 0.007218334 | 0.001604834 |
| | | Best | 1.59342E-09 | 4.15798E-11 | 0 | 1.30581E-08 | 0.00097117 |
| F38 | 0 | Mean | 4.70363E-05 | 1.39476E-06 | 1.78046E-06 | 3.14986E-05 | 0.000288005 |
| | | StdDev | 6.70904E-05 | 8.77149E-07 | 1.28089E-06 | 6.43525E-05 | 0.00013892 |
| | | Best | 3.91829E-08 | 6.72858E-11 | 4.82E-09 | 1.50435E-11 | 5.82234E-05 |
| F39 | -3.86 | Mean | -3.862782148 | -3.862782148 | -3.862782148 | -3.862782148 | -3.862782148 |
| | | StdDev | 2.65431E-15 | 2.68234E-15 | 2.69625E-15 | 2.71009E-15 | 2.71009E-15 |
| | | Best | -3.862782148 | -3.862782148 | -3.862782148 | -3.862782148 | -3.862782148 |
| F40 | -3.32 | Mean | -3.322368011 | -3.322368011 | -3.322368011 | -3.318394475 | -3.322368011 |
| | | StdDev | 6.54548E-16 | 5.71336E-16 | 5.14996E-16 | 0.021763955 | 6.54548E-16 |
| | | Best | -3.322368011 | -3.322368011 | -3.322368011 | -3.322368011 | -3.322368011 |
| F41 | 0 | Mean | 0 | 0 | 0.032798017 | 0.004761038 | 0 |
| | | StdDev | 0 | 0 | 0.018570459 | 0.008047673 | 0 |
| | | Best | 0 | 0 | 0.00739604 | 0 | 0 |
| F42 | 0 | Mean | 1.49806E-14 | 1.29674E-14 | 1.15463E-14 | 0.660186991 | 2.44545E-14 |
| | | StdDev | 3.29641E-15 | 3.31178E-15 | 3.61345E-15 | 0.711496752 | 3.02083E-15 |
| | | Best | 7.99361E-15 | 7.99361E-15 | 7.99361E-15 | 7.99361E-15 | 2.22045E-14 |
| F43 | 0 | Mean | 0 | 0 | 0.114662313 | 0.024187276 | 2.63417E-16 |
| | | StdDev | 0 | 0 | 0.532276418 | 0.080213839 | 0 |
| | | Best | 0 | 0 | 0 | 0 | 1.29727E-16 |
| F44 | 0 | Mean | 0 | 0 | 0 | 0 | 2.7797E-16 |
| | | StdDev | 0 | 0 | 0 | 0 | 0 |
| | | Best | 0 | 0 | 0 | 0 | 2.22214E-16 |
| F45 | -1.08 | Mean | -1.080938442 | -1.080938442 | -1.080938442 | -1.080938442 | -1.080938442 |
| | | StdDev | 4.70125E-16 | 4.51681E-16 | 4.51681E-16 | 4.51681E-16 | 4.96507E-16 |
| | | Best | -1.080938442 | -1.080938442 | -1.080938442 | -1.080938442 | -1.080938442 |
| F46 | -1.5 | Mean | -1.499999223 | -1.499999223 | -1.499999223 | -1.499999223 | -1.499999223 |
| | | StdDev | 6.77522E-16 | 6.77522E-16 | 6.77522E-16 | 6.77522E-16 | 1.05365E-15 |
| | | Best | -1.499999223 | -1.499999223 | -1.499999223 | -1.499999223 | -1.499999223 |
| F47 | NA | Mean | -1.403866666 | -1.320499999 | -1.271399999 | -1.069011938 | -1.482016588 |
| | | StdDev | 0.219139813 | 0.279253038 | 0.313658787 | 0.422205043 | 0.097662612 |
| | | Best | -1.5 | -1.5 | -1.5 | -1.5 | -1.499998488 |

| | | | | | | |
|---|---|---|---|---|---|---|
| F48 | Mean | 0 | 0 | 0 | 0 | 0 |
| | StdDev | 0 | 0 | 0 | 0 | 0 |
| | Best | 0 | 0 | 0 | 0 | 0 |
| F49 | Mean | 0 | 0 | 0 | 3.083487114 | 1.48707E-12 |
| | StdDev | 0 | 0 | 0 | 4.389694328 | 8.11041E-12 |
| | Best | 0 | 0 | 0 | 0 | 3.1715E-16 |
| F50 | Mean | 0 | 0 | 0 | 580.0839029 | 1.111095363 |
| | StdDev | 0 | 0 | 0 | 1280.698395 | 0.598962098 |
| | Best | 0 | 0 | 0 | 0 | 0.182153237 |

Table 3. Pair-wise statistical comparison of the MVS (m = 5, n = 50) algorithm by Wilcoxon Signed-Rank Test ($\alpha = 0.05$)

| Function | MVS (m = 5, n = 50) vs. VS | | | | MVS (m = 5, n = 50) vs. PSO2011 | | | | MVS (m = 5, n = 50) vs. ABC | | | |
|---|---|---|---|---|---|---|---|---|---|---|---|---|
| | p-value | T+ | T- | Winner | p-value | T+ | T- | winner | p-value | T+ | T- | winner |
| F1 | 1 | 0 | 0 | = | 1 | 0 | 0 | = | 1 | 0 | 0 | = |
| F2 | 0.014306 | 0 | 21 | + | 0.157299 | 0 | 3 | = | 1 | 0 | 0 | = |
| F3 | 1 | 0 | 0 | = | 1 | 0 | 0 | = | 1.73E-06 | 0 | 465 | + |
| F4 | 1 | 0 | 0 | = | 1 | 0 | 0 | = | 1.73E-06 | 0 | 465 | + |
| F5 | 0.000894 | 394 | 71 | - | 5.22E-06 | 454 | 11 | - | 1.73E-06 | 0 | 465 | + |
| F6 | 1 | 0 | 0 | = | 1 | 0 | 0 | = | 3.79E-06 | 0 | 406 | + |
| F7 | 1 | 0 | 0 | = | 1 | 0 | 0 | = | 1 | 0 | 0 | = |
| F8 | 1 | 0 | 0 | = | 1 | 0 | 0 | = | 0.067889 | 0 | 10 | = |
| F9 | 1 | 0 | 0 | = | 1 | 0 | 0 | = | 1.73E-06 | 0 | 465 | + |
| F10 | 0.012555 | 77 | 14 | - | 0.000311 | 91 | 0 | - | 1.4E-06 | 0 | 465 | + |
| F11 | 0.004964 | 209 | 44 | - | 3.32E-06 | 351 | 0 | - | 1.71E-06 | 0 | 465 | + |
| F12 | 1 | 0 | 0 | = | 1 | 0 | 0 | = | 1.73E-06 | 0 | 465 | + |
| F13 | 1.73E-06 | 0 | 465 | + | 1.92E-06 | 1 | 464 | + | 1.73E-06 | 0 | 465 | + |
| F14 | 0.179712 | 3 | 0 | = | 1.73E-06 | 0 | 465 | + | 0.000359 | 59 | 406 | + |
| F15 | 1 | 0 | 0 | = | 1 | 0 | 0 | = | 1.73E-06 | 0 | 465 | + |
| F16 | 1.73E-06 | 0 | 465 | + | 0.370935 | 276 | 189 | = | 1.73E-06 | 0 | 465 | + |
| F17 | 1.71E-06 | 0 | 465 | + | 1.58E-06 | 0 | 465 | + | 1.73E-06 | 0 | 465 | + |
| F18 | 0.317311 | 1 | 0 | = | 0.140773 | 1 | 9 | = | 1.73E-06 | 0 | 465 | + |
| F19 | 1 | 0 | 0 | = | 1 | 0 | 0 | = | 1 | 0 | 0 | = |
| F20 | 1 | 0 | 0 | = | 1 | 0 | 0 | = | 1 | 0 | 0 | = |
| F21 | 1 | 0 | 0 | = | 1 | 0 | 0 | = | 1 | 0 | 0 | = |
| F22 | 1.73E-06 | 0 | 465 | + | 1.72E-06 | 0 | 465 | + | 0.019631 | 21 | 0 | - |
| F23 | 1.73E-06 | 0 | 465 | + | 1.73E-06 | 0 | 465 | + | 1.08E-05 | 253 | 0 | - |
| F24 | 1 | 0 | 0 | = | 1 | 0 | 0 | = | 1 | 0 | 0 | = |
| F25 | 0.151957 | 215 | 110 | = | 0.005349 | 247 | 53 | - | 3.37E-06 | 406 | 0 | - |
| F26 | 0.03001 | 127 | 338 | + | 0.015658 | 350 | 115 | - | 1.73E-06 | 465 | 0 | - |
| F27 | 1 | 0 | 0 | = | 1 | 0 | 0 | = | 1 | 0 | 0 | = |
| F28 | 1 | 0 | 0 | = | 0.317311 | 0 | 1 | = | 1 | 0 | 0 | = |
| F29 | 1 | 0 | 0 | = | 1 | 0 | 0 | = | 1 | 0 | 0 | = |

Table 3. (continued).

| Function | MVS (m = 5, n = 50) vs. VS | | | | MVS (m = 5, n = 50) vs. PSO2011 | | | | MVS (m = 5, n = 50) vs. ABC | | | |
|---|---|---|---|---|---|---|---|---|---|---|---|---|
| | p-value | T+ | T- | winner | p-value | T+ | T- | winner | p-value | T+ | T- | winner |
| F32 | 0.001486 | 173 | 17 | - | 2.84E-05 | 271 | 5 | - | 0.403741 | 127 | 83 | = |
| F33 | 0.078122 | 261.5 | 116.5 | = | 1.22E-05 | 370.5 | 7.5 | - | 1.73E-06 | 0 | 465 | + |
| F34 | 0.008151 | 28 | 0 | - | 0.647219 | 28 | 38 | = | 0.008151 | 28 | 0 | - |
| F35 | 0.365712 | 42 | 24 | = | 0.0027 | 45 | 0 | - | 0.002282 | 7 | 84 | + |
| F36 | 0.011617 | 112 | 24 | - | 8.85E-07 | 378 | 0 | - | 0.365712 | 24 | 42 | = |
| F37 | 0.280214 | 285 | 180 | = | 0.065641 | 322 | 143 | = | 0.861213 | 224 | 241 | = |
| F38 | 9.32E-06 | 448 | 17 | - | 0.054463 | 326 | 139 | = | 2.13E-06 | 2 | 463 | + |
| F39 | 0.179712 | 12 | 3 | = | 0.0455 | 10 | 0 | - | 0.0455 | 10 | 0 | - |
| F40 | 0.004678 | 36 | 0 | - | 0.020137 | 66 | 12 | - | 1 | 68 | 68 | = |
| F41 | 1.73E-06 | 0 | 465 | + | 0.005683 | 6 | 85 | + | 0.0455 | 10 | 0 | - |
| F42 | 0.000607 | 143 | 10 | - | 0.014133 | 105 | 330 | + | 3.46E-06 | 0 | 378 | + |
| F43 | 0.042168 | 0 | 15 | + | 0.0656 | 0 | 10 | = | 1.73E-06 | 0 | 465 | + |
| F44 | 1 | 0 | 0 | = | 1 | 0 | 0 | = | 1.73E-06 | 0 | 465 | + |
| F45 | 0.157299 | 3 | 0 | = | 0.157299 | 3 | 0 | = | 0.256839 | 8 | 20 | = |
| F46 | 1 | 0 | 0 | = | 1 | 0 | 0 | = | 8.83E-07 | 0 | 435 | + |
| F47 | 0.030533 | 11.5 | 66.5 | + | 0.001812 | 17.5 | 172.5 | + | 0.051931 | 138 | 327 | = |
| F48 | 1 | 0 | 0 | = | 1 | 0 | 0 | = | 1 | 0 | 0 | = |
| F49 | 1 | 0 | 0 | = | 0.003346 | 0 | 66 | + | 1.73E-06 | 0 | 465 | + |
| F50 | 1 | 0 | 0 | = | 8.84E-05 | 0 | 210 | + | 1.73E-06 | 0 | 465 | + |
| +/=/- | 10/31/9 | | | | 11/28/11 | | | | 24/18/8 | | | |

International Journal of Artificial Intelligence and Applications (IJAIA), Vol. 7, No. 3, May 2016

Table 4. Pair-wise statistical comparison of the MVS (m = 5, n = 250) algorithm by Wilcoxon Signed-Rank Test ($\alpha = 0.05$).

| Function | MVS (m = 5, n = 250) vs. VS | | | | MVS (m = 5, n = 250) vs. PSO2011 | | | | MVS (m = 5, n = 250) vs. ABC | | | |
|---|---|---|---|---|---|---|---|---|---|---|---|---|
| | p-value | T+ | T- | Winner | p-value | T+ | T- | winner | p-value | T+ | T- | winner |
| F1 | 1 | 0 | 0 | = | 1 | 0 | 0 | = | 1 | 0 | 0 | = |
| F2 | 0.014306 | 0 | 21 | + | 0.157299 | 0 | 3 | = | 1 | 0 | 0 | = |
| F3 | 1 | 0 | 0 | = | 1 | 0 | 0 | = | 1.73E-06 | 0 | 465 | + |
| F4 | 1 | 0 | 0 | = | 1 | 0 | 0 | = | 1.73E-06 | 0 | 465 | + |
| F5 | 0.289477 | 181 | 284 | = | 2.35E-06 | 462 | 3 | - | 1.73E-06 | 0 | 465 | + |
| F6 | 1 | 0 | 0 | = | 1 | 0 | 0 | = | 3.79E-06 | 0 | 406 | + |
| F7 | 1 | 0 | 0 | = | 1 | 0 | 0 | = | 1 | 0 | 0 | = |
| F8 | 1 | 0 | 0 | = | 1 | 0 | 0 | = | 0.067889 | 0 | 10 | = |
| F9 | 1 | 0 | 0 | = | 1 | 0 | 0 | = | 1.73E-06 | 0 | 465 | + |
| F10 | 0.4795 | 22.5 | 13.5 | = | 0.014306 | 21 | 0 | - | 1.2E-06 | 0 | 465 | + |
| F11 | 0.226741 | 80 | 40 | = | 7.9E-05 | 171 | 0 | - | 1.7E-06 | 0 | 465 | + |
| F12 | 1 | 0 | 0 | = | 1 | 0 | 0 | = | 1.73E-06 | 0 | 465 | + |
| F13 | 1.73E-06 | 0 | 465 | + | 1.73E-06 | 0 | 465 | + | 1.73E-06 | 0 | 465 | + |
| F14 | 1 | 0 | 0 | = | 1.73E-06 | 0 | 465 | + | 1.73E-06 | 0 | 465 | + |
| F15 | 1 | 0 | 0 | = | 1 | 0 | 0 | = | 1.73E-06 | 0 | 465 | + |
| F16 | 1.73E-06 | 0 | 465 | + | 0.370935 | 276 | 189 | = | 1.73E-06 | 0 | 465 | + |
| F17 | 1.71E-06 | 0 | 465 | + | 1.58E-06 | 0 | 465 | + | 1.73E-06 | 0 | 465 | + |
| F18 | 1 | 0 | 0 | = | 0.10247 | 0 | 6 | = | 1.73E-06 | 0 | 465 | + |
| F19 | 1 | 0 | 0 | = | 1 | 0 | 0 | = | 1 | 0 | 0 | = |
| F20 | 1 | 0 | 0 | = | 1 | 0 | 0 | = | 1 | 0 | 0 | = |
| F21 | 1 | 0 | 0 | = | 1 | 0 | 0 | = | 1 | 0 | 0 | = |
| F22 | 1.73E-06 | 0 | 465 | + | 1.72E-06 | 0 | 465 | + | 0.317311 | 1 | 0 | = |
| F23 | 1.73E-06 | 0 | 465 | + | 1.73E-06 | 0 | 465 | + | 0.000162 | 120 | 0 | - |
| F24 | 1 | 0 | 0 | = | 1 | 0 | 0 | = | 1 | 0 | 0 | = |
| F25 | 0.369058 | 117 | 73 | = | 0.005357 | 229 | 47 | + | 1.4E-06 | 465 | 0 | + |
| F26 | 0.006424 | 100 | 365 | + | 0.093676 | 314 | 151 | = | 1.73E-06 | 465 | 0 | + |
| F27 | 1 | 0 | 0 | = | 1 | 0 | 0 | = | 1 | 0 | 0 | = |
| F28 | 1 | 0 | 0 | = | 0.317311 | 0 | 1 | = | 1 | 0 | 0 | = |
| F29 | 1 | 0 | 0 | = | 1 | 0 | 0 | = | 1 | 0 | 0 | = |
| F30 | 1 | 0 | 0 | = | 1 | 0 | 0 | = | 0.001766 | 0 | 66 | + |
| F31 | 0.039566 | 64.5 | 188.5 | + | 6.01E-06 | 0 | 351 | + | 0.040712 | 199 | 77 | - |

Table 4. (continued).

| Function | MVS (m = 5, n = 250) vs. VS | | | | MVS (m = 5, n = 250) vs. PSO2011 | | | | MVS (m = 5, n = 250) vs. ABC | | | |
|---|---|---|---|---|---|---|---|---|---|---|---|---|
| | p-value | T+ | T- | winner | p-value | T+ | T- | winner | p-value | T+ | T- | winner |
| F32 | 0.552741 | 99 | 72 | = | 0.011249 | 172 | 38 | - | 0.017838 | 48 | 183 | + |
| F33 | 0.827955 | 167 | 184 | = | 0.002577 | 356 | 79 | - | 1.73E-06 | 0 | 465 | + |
| F34 | 0.0455 | 10 | 0 | - | 0.256108 | 10 | 26 | = | 0.0455 | 10 | 0 | - |
| F35 | 0.738883 | 20 | 25 | = | 0.025347 | 15 | 0 | - | 0.000579 | 20 | 170 | + |
| F36 | 0.157299 | 27 | 9 | = | 9.63E-07 | 300 | 0 | - | 0.003892 | 0 | 45 | + |
| F37 | 0.452807 | 269 | 196 | = | 0.059836 | 324 | 141 | - | 0.893644 | 226 | 239 | = |
| F38 | 0.125438 | 158 | 307 | = | 0.000616 | 66 | 399 | + | 1.73E-06 | 0 | 465 | + |
| F39 | 0.563703 | 4 | 2 | = | 0.157299 | 3 | 0 | = | 0.157299 | 3 | 0 | = |
| F40 | 0.317311 | 30 | 15 | = | 0.205903 | 21 | 7 | = | 0.165518 | 28 | 63 | = |
| F41 | 1.73E-06 | 0 | 465 | + | 0.005005 | 0 | 55 | + | 1 | 0 | 0 | = |
| F42 | 0.05778 | 44 | 11 | = | 0.007028 | 78 | 300 | + | 1.24E-06 | 0 | 465 | + |
| F43 | 0.042168 | 0 | 15 | + | 0.0656 | 0 | 10 | = | 1.73E-06 | 0 | 465 | + |
| F44 | 1 | 0 | 0 | = | 1 | 0 | 0 | = | 1.73E-06 | 0 | 465 | + |
| F45 | 1 | 0 | 0 | = | 1 | 0 | 0 | = | 0.025347 | 0 | 15 | + |
| F46 | 1 | 0 | 0 | = | 1 | 0 | 0 | = | 8.83E-07 | 0 | 435 | + |
| F47 | 0.567938 | 57 | 79 | = | 0.0033 | 22 | 168 | + | 0.975387 | 234 | 231 | = |
| F48 | 1 | 0 | 0 | = | 1 | 0 | 0 | = | 1 | 0 | 0 | = |
| F49 | 1 | 0 | 0 | = | 0.003346 | 0 | 66 | + | 1.73E-06 | 0 | 465 | + |
| F50 | 1 | 0 | 0 | = | 8.84E-05 | 0 | 210 | + | 1.73E-06 | 0 | 465 | + |
| +/=/- | 10/39/1 | | | | 13/29/8 | | | | 29/18/3 | | | |

As shown in Table 4, MVS algorithm with 250 candidate solutions can successfully improve the near optimal solutions and thus performs better than all of the other algorithms.

In [31], authors stated that after a sufficient value for colony size, any increment in the value does not improve the performance of the ABC algorithm significantly. For the test problems carried out in [31] colony sizes of 10, 50 and 100 are used for the ABC algorithm. It is shown

International Journal of Artificial Intelligence and Applications (IJAIA), Vol. 7, No. 3, May 2016

that although from 10 to 50 the performance of the ABC algorithm significantly increased, there is not any significant difference between the performances achieved by 50 and 100 colony sizes. Similarly, for the PSO algorithm it is reported that, PSO with different population sizes has almost the similar performance which means the performance of PSO is not sensitive to the population size [32]. Based on the above considerations, in this study a comparison of the MVS algorithm to the ABC and PSO2011 algorithms with a different population size is not performed. For the VS algorithm it is expected to achieve better exploitation ability with an increased number of candidate solutions. But the problem with the VS algorithm is with its global search ability rather than the local search ability for some of the functions listed above. Therefore, a comparison of the MVS (m = 5, n = 50) to VS algorithm with 50 candidate solutions is thought to be enough to show the improvement achieved by the modification performed on the VS algorithm.

In Figure 7, the average computational time of 30 runs for 500,000 iterations is also provided for the MVS (m = 5, n = 50), MVS (m = 5, n = 250), VS, PSO2011 and ABC algorithms. As shown in this figure, the required computational time to perform 500,000 iterations with the MVS algorithm is slightly increased when compared to the VS algorithm. However, even for the MVS (m = 5, n = 250) algorithm the required computational time to perform 500,000 iterations is still lower than the PSO2011 and ABC algorithms.

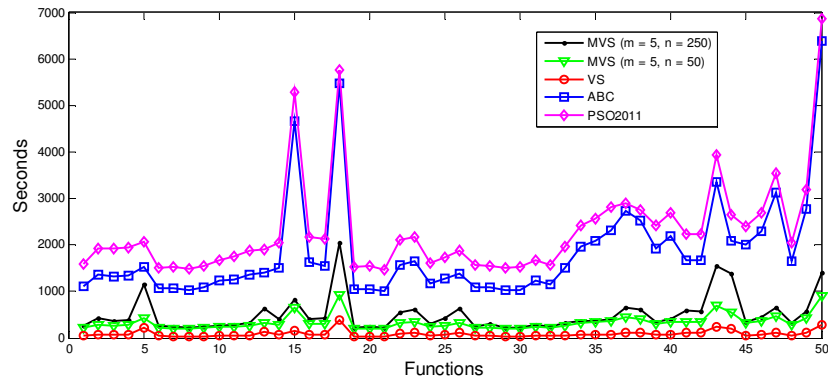

Figure 7. Average computational time of 30 runs for 50 benchmark functions (500,000 iterations).

## 4. CONCLUSIONS

This paper presents a modified VS algorithm in which the global search ability of the existing VS algorithm is improved. This is achieved by using multiple centers during the candidate solution generation phase of the algorithm at each iteration pass. In the VS algorithm, only one center is used for this purpose and this usually leads the algorithm to being trapped into a local minimum point for some of the benchmark functions. Although the complexity of the existing VS algorithm is a bit increased, there is not any significant difference between the computational time of the modified VS algorithm and the existing VS algorithm. Computational results showed that the MVS algorithm outperforms the existing VS algorithm, PSO2011 and ABC algorithms for the benchmark numerical function set.

In the future studies, the MVS algorithm will be used for the solution of some real world optimization problems.

**Authors**

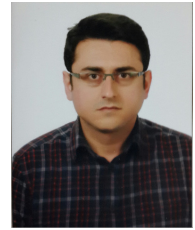

Dr. Berat Doğan received his BSc. degree in Electronics Engineering from Erciyes University, Turkey, 2006. He received his MSc. degree in Biomedical Engineering from Istanbul Technical University, Turkey, 2009. He received his PhD. in Electronics Engineering at Istanbul Technical University, Turkey, 2015. Between 2008-2009 he worked as a software engineer at Nortel Networks Netas Telecommunication Inc. Then, from 2009 to July 2015 he worked as a Research Assistant at Istanbul Technical University. Now he is working as an Assistant Professor at Inonu University, Malatya, Turkey. His research interests include optimization algorithms, pattern recognition, biomedical signal and image processing, and bioinformatics.